\documentclass{article}
\usepackage[main, final]{neurips_2026}
\usepackage[table]{xcolor}

\usepackage[utf8]{inputenc} 
\usepackage[T1]{fontenc}    
\usepackage{hyperref}       
\usepackage{url}            
\usepackage{booktabs}       
\usepackage{amsfonts}       
\usepackage{nicefrac}       
\usepackage{microtype}      
\usepackage{xcolor}         
\usepackage{amsmath}
\usepackage{graphicx}
\usepackage{wrapfig}
\usepackage{booktabs}
\usepackage[table]{xcolor}
\usepackage{caption}
\usepackage{subcaption}

\title{Exact Flow Linear Attention: Exact Solution from Continuous-Time Dynamics}

\author{
  Jingdi Lei$^{1*}$,\enspace 
  Di Zhang$^{2*}$,\enspace 
  \textbf{Soujanya Poria}$^{1}$\thanks{Corresponding author} \enspace
  \\
  \textsuperscript{1}\small{Nanyang Technological University},\enspace
  \textsuperscript{2}\small{Fudan University}\\
  \small{\texttt{jingdi001@e.ntu.edu.sg}}, \quad \small{\texttt{di.zhang@ustc.edu}}, \\
  \small{\texttt{soujanya.poria@ntu.edu.sg}}
  \vspace{-5.mm}
}

\begin{document}

\maketitle

\begin{abstract}
In this paper, we introduce \textbf{Exact Flow Linear Attention~(EFLA)}, an exact-flow formulation of delta-rule linear attention. We show that the delta-rule update can be interpreted as an explicit Euler discretization of an underlying continuous-time system. EFLA replaces this first-order update with the exact closed-form flow. By exploiting the \textit{rank-1} structure of the dynamics matrix, both the matrix exponential and the input integral collapse to a simple update that preserves delta-rule linear attention's algebraic structure, parameter count, linear-time complexity, and chunkwise parallelism. This attention mechanism removes the Euler discretization error of the delta-rule dynamics without introducing additional parameters. Experiments on robustness tests, language modeling benchmarks, and the MAD synthetic benchmark show that EFLA improves stability under corrupted and high-energy inputs, reduces perplexity, and achieves stronger downstream performance compared to SSM and Euler-style baselines. These results establish exact-flow integration as a principled and scalable update mechanism for delta-rule linear attention. Code is available at:
\href{https://github.com/declare-lab/EFLA}{https://github.com/declare-lab/EFLA}.
\end{abstract}

\section{Introduction}
\label{sec:introduction}

As large language models (LLMs) evolve into increasingly capable agents~\citep{yao2022react, team2025tongyi, google_gemini_deep_research, openai_deep_research_intro}, inference efficiency has become a critical bottleneck~\citep{dao2022flashattention, kwon2023efficient, kim2024llm}. This challenge is especially pronounced in long-context processing and reinforcement learning (RL) environments~\citep{guo2025deepseek, lai2025survey}, where models are required to process extended reasoning trajectories, maintain long-range dependencies, and engage in complex tool-use interactions~\citep{lightman2023let, yao2023tree}. In these scenarios, the quadratic time complexity of standard softmax attention~\citep{vaswani2017attention} introduces substantial computational overhead, constraining model throughput, scalability to long
contexts, and real-time interactivity.~\citep{liu2023ring, jiang2024longllmlingua, katharopoulos2020transformers}.

This has motivated a broad line of work on linear-time sequence models, including linear attention and state space models~(SSMs). Among them, delta-rule linear attention~\citep{schlag2021linear, yang2024parallelizing} provides an appealing recurrent formulation by maintaining a matrix-valued associative memory. Compared with purely additive linear attention~\citep{katharopoulos2020transformers}, the delta rule introduces an input-dependent correction mechanism: the memory is updated by partially erasing the component associated with the current key and injecting a new key--value association. This update admits efficient recurrent and chunkwise-parallel implementations~\citep{yang2024parallelizing}. Despite its empirical effectiveness, this update is usually motivated as a discrete online learning rule, leaving its underlying continuous-time structure less explored.

In this work, we revisit delta-rule linear attention from the perspective of numerical integration. Under a zero-order-hold~(ZOH) assumption, where the current key and value are treated as fixed within each token interval, the delta-rule update can be interpreted as an explicit Euler discretization of a continuous-time system. This interpretation exposes a principled source of approximation error: the Euler update is only first-order accurate. Although simple and efficient, Euler discretization can introduce accumulated discretization error and may become inaccurate when the effective dynamics are stiff, for example under large key norms or large update scales. Existing methods often mitigate such issues by introducing gating mechanisms including decay factors, gating functions~\citep{dao2024transformers, ma2022mega, sun2023retentive}, or adaptive forgetting coefficients~\citep{yang2023gated, yang2024gated, team2025kimi}. This motivates a natural question: instead of heuristically mitigating the instability of Euler-style updates, can we derive an exact update directly from the underlying continuous-time dynamics of delta-rule linear attention?

We answer this question by proposing \textbf{Exact Flow Linear Attention~(EFLA)}, a drop-in replacement for the delta-rule update that solves the corresponding ZOH continuous-time dynamics in closed form. From the first principle, we eliminate the discretization errors by solving the underlying ODE exactly. EFLA can be mathematically interpreted as the general solution of the first-order ODE, yielding a continuous-time, exact flow update of linear attention. The key observation is that the transition matrix is rank-1 in delta-rule update. Therefore, although a generic matrix exponential would be computationally expensive, both the exponential transition and the input integral collapse into a simple closed form. EFLA preserves the algebraic structure and computational order of delta-rule linear attention while removing the Euler discretization error of the underlying ZOH dynamics.

Since EFLA retains the same rank-1 update form as vanilla delta-rule linear attention, it can be integrated with existing hardware-efficient WY/UT-based chunkwise parallelization schemes. This allows EFLA to preserve linear-time recurrent inference and efficient parallel training. Empirically, we evaluate EFLA on robustness tests, language modeling benchmarks, and the MAD synthetic benchmark. Across these evaluations, EFLA consistently improves over previous Euler discretization baselines in both robustness, convergence, and downstream performance while maintaining comparable training throughput and introducing no additional parameters.

Our contributions are summarized as follows:
\begin{itemize}
    \item We reinterpret delta-rule linear attention as an explicit Euler discretization of an underlying ZOH continuous-time ODE, and propose \textbf{Exact Flow Linear Attention~(EFLA)}, which solves this ODE in closed form.

    \item We show that the rank-1 structure of delta-rule learning makes the exact matrix exponential and input integral analytically tractable, yielding an exact-flow update with the same computational order and algebraic form as delta-rule linear attention.

    \item We validate EFLA across robustness tests, language modeling benchmarks, and the MAD synthetic benchmark, showing consistent improvements over previous Euler-style baselines while maintaining comparable throughput.
\end{itemize}

\section{Preliminary}
\label{sec:background}

\textbf{Scaled Dot-Product Attention.}
Given queries \(\mathbf{Q}\in\mathbb{R}^{n\times d}\), keys
\(\mathbf{K}\in\mathbb{R}^{n\times d}\), and values
\(\mathbf{V}\in\mathbb{R}^{n\times d}\), scaled dot-product attention~\citep{vaswani2017attention}
is defined as:
\begin{equation}
\mathrm{Attn}(\mathbf{Q},\mathbf{K},\mathbf{V})
=
\mathrm{softmax}
\left(
\frac{\mathbf{Q}\mathbf{K}^{\top}}{\sqrt{d}}+\mathbf{M}
\right)
\mathbf{V},
\end{equation}
where \(\mathbf{M}\in\mathbb{R}^{n\times n}\) is the additive causal mask,
with zeros on and below the diagonal and \(-\infty\) above.

\textbf{Linear Attention as Associative Memory.}
Linear attention~\citep{katharopoulos2020transformers} maintains a
matrix-valued recurrent state that accumulates key--value associations:
\begin{equation}
\mathbf{S}_t
=
\mathbf{S}_{t-1}
+
\mathbf{k}_t\mathbf{v}_t^\top,
\qquad
\mathbf{o}_t
=
\mathbf{S}_t^\top\mathbf{q}_t .
\end{equation}
From the fast-weight perspective~\citep{schlag2021linear},
\(\mathbf{S}_t\) serves as an associative memory that stores transient mappings
from keys to values. This additive update can be viewed as gradient descent on
the unbounded correlation objective:
\begin{equation}
\mathcal{L}_t(\mathbf{S})
=
-
\langle
\mathbf{S}^\top\mathbf{k}_t,
\mathbf{v}_t
\rangle .
\end{equation}
Since this objective only reinforces new key--value pairs and does not provide
an explicit mechanism for erasing old associations, the accumulated memory may
grow without bound and suffer from interference over long contexts.

\textbf{Delta-rule Learning.}
Delta-rule linear attention addresses this issue by formulating the memory state
as the parameter of an online reconstruction problem:
\begin{equation}
\mathcal{L}_t(\mathbf{S})
=
\frac{1}{2}
\left\lVert
\mathbf{S}^\top\mathbf{k}_t-\mathbf{v}_t
\right\rVert^2 .
\end{equation}
Taking one gradient step with learning rate \(\beta_t\) gives:
\begin{equation}
\begin{aligned}
\mathbf{S}_t
&=
\mathbf{S}_{t-1}
-
\beta_t
\nabla_{\mathbf{S}}
\mathcal{L}_t(\mathbf{S}_{t-1})
=
\left(
\mathbf{I}
-
\beta_t\mathbf{k}_t\mathbf{k}_t^\top
\right)
\mathbf{S}_{t-1}
+
\beta_t\mathbf{k}_t\mathbf{v}_t^\top .
\end{aligned}
\end{equation}
This update corrects the associative memory toward the mapping
\(\mathbf{k}_t\mapsto\mathbf{v}_t\). Its rank-1 correction structure enables
WY-style products of rank-1 transformations, facilitating hardware-efficient
chunkwise parallelization~\citep{bischof1987wy, yang2024parallelizing}.

\textbf{Euler Methods in ODEs.}
Given a first-order ordinary differential equation~(ODE):
\begin{equation}
\frac{d\mathbf{S}(\tau)}{d\tau}
=
f(\tau,\mathbf{S}(\tau)),
\end{equation}
numerical integration methods approximate the solution at discrete time points.
The explicit Euler method~\citep{euler1792institutiones} updates the state by
using the derivative at the current position:
\begin{equation}
\mathbf{S}_{t}
=
\mathbf{S}_{t-1}
+
\beta_t
f(t-1,\mathbf{S}_{t-1}),
\end{equation}
where \(\beta_t\) is the step size. Euler discretization is computationally
simple but only first-order accurate, with local truncation error
\(\mathcal{O}(\beta_t^2)\). As shown below, Euler discretization delta-rule updates
can be interpreted as such an explicit Euler discretization of an underlying
continuous-time memory dynamics.

\section{Method}
\label{sec:efla}

\subsection{Exact Flow of Delta-rule Linear Attention}
\label{sec:exact_flow_delta_rule}
We begin by revisiting DeltaNet~\citep{schlag2021linear}, which formulates linear attention as online gradient descent on a reconstruction objective:
\begin{equation}
    \mathcal{L}_t(\mathbf{S}) = \frac{1}{2} \| \mathbf{S}^\top \mathbf{k}_t - \mathbf{v}_t \|^2.
\end{equation}

Applying a single gradient descent step with learning rate $\beta_t$ yields:

\begin{equation}
    \mathbf{S}_t 
= \mathbf{S}_{t-1} + \beta_t \left( -\mathbf{k}_t \mathbf{k}_t^\top \mathbf{S}_{t-1} + \mathbf{k}_t \mathbf{v}_t^\top \right).
\end{equation}

To formalize the underlying dynamics, we define the dynamics matrix
$\mathbf{A}_t = \mathbf{k}_t\mathbf{k}_t^{\top}$ and the input forcing term
$\mathbf{b}_t = \mathbf{k}_t\mathbf{v}_t^{\top}$. Since the input sequence is
observed only at discrete time steps, we adopt the Zero-Order Hold
(ZOH)~\citep{iserles2009first} assumption to construct a continuous-time interpolation. Under this assumption, $\mathbf{A}_t$ and $\mathbf{b}_t$ are
treated as piecewise constant within each update interval. The system evolves according to a first-order ODE:
\begin{equation}
\label{eq:general_solution_ode}
    \frac{d\mathbf{S}(t)}{dt}=-\mathbf{A}_t \mathbf{S}(t) + \mathbf{b}_t .
\end{equation}

It reveals that the standard Delta rule update is only a first-order numerical approximation to the underlying continuous dynamics. This observation motivates a natural question: can we avoid the Euler discretization error by solving the ODE exactly? Eq.~\eqref{eq:general_solution_ode} admits a closed-form matrix-exponential solution, yielding\footnote{The detailed derivation of this closed-form expression is provided in Appendix~\ref{appendix:general-solution-ode}.}:
\begin{equation}
    \mathbf{S}_{t}
    =
    e^{-\beta_t\mathbf{A}_t}\mathbf{S}_{t-1}
    +
    \int_0^{\beta_t}
    e^{-(\beta_t - \tau)\mathbf{A}_t}
    \mathbf{b}_t
    \, d\tau .
\end{equation}

\subsection{Efficient Exact Flow via the \textit{Rank-1} Structure}

While the infinite-order solution eliminates discretization error, naively evaluating the matrix exponential $e^{-\beta_t \mathbf{A}_t}$ for a general matrix typically requires $\mathcal{O}(d^3)$ complexity~\citep{gu2020hipporecurrentmemoryoptimal}. We bypass this computational bottleneck by leveraging the \textit{rank-1} structure of the dynamics matrix $\mathbf{A}_t = \mathbf{k}_t \mathbf{k}_t^\top$, which allows the exponential to be computed in linear time. 

We observe that $\mathbf{A}_t$ satisfies the idempotence-like property $\mathbf{A}_t^n = \lambda_t^{n-1} \mathbf{A}_t$ for $n \ge 1$, where $\lambda_t = \mathbf{k}_t^\top \mathbf{k}_t$ (see Appendix~\ref{appendix:property-rank1-matrices} for the proof).
This property allows us to collapse the Taylor series of the matrix exponential into a computable closed form:
\begin{equation}
e^{-\beta_t \mathbf{A}_t} 
= \mathbf{I} + \sum_{n=1}^{\infty} \frac{(-\beta_t)^n}{n!} \mathbf{A}_t^n 
= \mathbf{I} - \frac{1 - e^{-\beta_t \lambda_t}}{\lambda_t} \mathbf{A}_t.
\end{equation}
Substituting this transition operator into the integral term $\int_0^{\beta_t} e^{-(\beta_t - \tau)\mathbf{A}_t} \mathbf{b}_t \, d\tau$ yields the exact input injection:

\begin{equation}
\begin{aligned}
\mathbf{I}_t &= \int_0^{\beta_t} \left( \mathbf{I} - \frac{1 - e^{-\lambda_t (\beta_t - \tau)}}{\lambda_t} \mathbf{A}_t \right) \mathbf{b}_t \, d\tau \\
&= \beta_t \mathbf{b}_t - \frac{\mathbf{A}_t \mathbf{b}_t}{\lambda_t} \left( \beta_t - \frac{1 - e^{-\beta_t \lambda_t}}{\lambda_t} \right).
\end{aligned}
\end{equation}

Crucially, since $\mathbf{b}_t = \mathbf{k}_t \mathbf{v}_t^\top$ and $\mathbf{A}_t = \mathbf{k}_t \mathbf{k}_t^\top$, we have $\mathbf{A}_t \mathbf{b}_t = \lambda_t \mathbf{b}_t$. This algebraic relationship allows for significant simplification of the integral term:

\begin{equation}
    \mathbf{I}_t = \beta_t \mathbf{b}_t - \beta_t \mathbf{b}_t + \frac{1 - e^{-\beta_t \lambda_t}}{\lambda_t} \mathbf{b}_t = \frac{1 - e^{-\beta_t \lambda_t}}{\lambda_t} \mathbf{b}_t.
\end{equation}

Combining these results, the final Exact Flow Linear Attention update rule is given by:
\begin{equation} \label{eq:efla_final}
\mathbf{S}_t = \left(\mathbf{I} - \frac{1 - e^{-\beta_t \lambda_t}}{\lambda_t} \mathbf{k}_t\mathbf{k}_t^{\top}\right)\mathbf{S}_{t-1} + \frac{1 - e^{-\beta_t \lambda_t}}{\lambda_t} \mathbf{k}_t\mathbf{v}_t^{\top}.
\end{equation}

Thus, EFLA has the same rank-1 algebraic form as the delta-rule update, with
\(\alpha_t\) replacing \(\beta_t\). It removes the Euler discretization error of
the ZOH dynamics while preserving linear complexity in the sequence length,
namely \(\mathcal{O}(Ld^2)\) for a length-\(L\) sequence with a \(d\times d\)
memory state.

\subsection{Chunkwise Parallelism Form}
\label{sec:chunkwise_parallelism}
We observe that the EFLA update rule shares an identical algebraic structure with DeltaNet. Given this structural equivalence, we can seamlessly adapt the hardware-efficient parallelization strategies originally developed for DeltaNet~\citep{yang2024parallelizing}. In this section, we derive the chunkwise parallel formulation specifically for EFLA.

To derive the chunkwise parallel form, we first unroll the recurrence relation. Denoting $\frac{1 - e^{-\beta_t \lambda_t}}{\lambda_t} = \alpha_t$, the state update becomes:

\begin{equation}
\begin{aligned}
\mathbf{S}_t 
&= (\mathbf{I} - \alpha_t \mathbf{k}_t \mathbf{k}_t^{\top})\mathbf{S}_{t-1}
   + \alpha_t \mathbf{k}_t\mathbf{v}_t^{\top} = \sum_{i=1}^t 
   \left(
   \prod_{j=i+1}^t  
   (\mathbf{I}- \alpha_j \mathbf{k}_j \mathbf{k}_j^{\top})
   \right)
   \alpha_i (\mathbf{k}_i\mathbf{v}_i^{\top}).
\end{aligned}
\end{equation}

Then we can define the following variables:
\begin{equation}
\mathbf{P}_i^j = \prod_{t=i}^{j}(\mathbf{I}-\alpha_t \mathbf{k}_t \mathbf{k}_t^\top),
\qquad
\mathbf{H}_i^j =
\sum_{t=i}^{j}\mathbf{P}_{t+1}^j
\alpha_t \mathbf{k}_t \mathbf{v}_t^\top
\end{equation}
where $\mathbf{P}_i^j=\mathbf{I}$ when $i>j$. $\mathbf{P}_i^j$ can be considered as decay factor applied to $\mathbf{S}_i$ to obtain $\mathbf{S}_j$ and $\mathbf{H}_i^j$ is an accumulation term to $\mathbf{S}_j$ from token $i$. The Chunkwise can be written as follows:

\begin{equation}
    \mathbf{S}_{[t]}^{r} = \mathbf{P}_{[t]}^r \mathbf{S}_{[t]}^0 + \mathbf{H}_{[t]}^r
\end{equation}

where we define the chunkwise variables $\mathbf{S}^{r}_{[t]} = \mathbf{S}_{t C + r}$, $\mathbf{P}^{r}_{[t]} = \mathbf{P}^{t C + r}_{t C + 1}$, 
$\mathbf{H}^{r}_{[t]} = \mathbf{H}^{t C + r}_{t C + 1}$. Here we have $\frac{L}{C}$ chunks of size $C$. We can use induction to derive the WY representations of $\mathbf{P}^{r}_{[t]}$ and $\mathbf{H}^{r}_{[t]}$:

\begin{equation}
    \mathbf{P}_{[t]}^r=\mathbf{I}-\sum_{i=1}^r  \mathbf{k}_{[t]}^{i}\mathbf{w}_{[t]}^{i \top},
\qquad
\mathbf{H}_{[t]}^r=\sum_{i=1}^r \mathbf{k}_{[t]}^i\mathbf{u}_{[t]}^{i\top}
\end{equation}

\begin{equation}
\mathbf{w}_{[t]}^{r\top}=\alpha_{[t]}^{r}\left(\mathbf{k}_{[t]}^{r\top}-\sum_{i=1}^{r-1}(\mathbf{k}_{[t]}^{r \top}\mathbf{k}_{[t]}^i)\mathbf{w}_{[t]}^{i \top}\right),
\quad
\mathbf{u}_{[t]}^{r\top}=\alpha_{[t]}^r\left(\mathbf{v}_{[t]}^{r\top}-\sum_{i=1}^{r-1}(\mathbf{k}_{[t]}^{r \top}\mathbf{k}_{[t]}^i)\mathbf{u}_{[t]}^{i\top}\right).
\end{equation}

subsequently, we can obtain the chunk-level recurrence for states and outputs:

\begin{equation}
\begin{aligned}
\mathbf{S}_{[t]}^r
&= \mathbf{S}_{[t]}^0
 - \left(
   \sum_{i=1}^r
   \mathbf{k}_{[t]}^{i}\mathbf{w}_{[t]}^{i \top}
   \right)
   \mathbf{S}_{[t]}^0
 + \sum_{i=1}^r
   \mathbf{k}_{[t]}^i\mathbf{u}_{[t]}^{i\top} = \mathbf{S}_{[t]}^0
 + \sum_{i=1}^r
   \mathbf{k}_{[t]}^i
   \bigl(
   \mathbf{u}_{[t]}^{i\top}
   - \mathbf{w}_{[t]}^{i \top}\mathbf{S}_{[t]}^0
   \bigr).
\end{aligned}
\end{equation}

\begin{equation}
\mathbf{o}_{[t]}^r=\mathbf{S}_{[t]}^{r \top}\mathbf{q}_{[t]}^r=\mathbf{S}_{[t]}^{0 \top}\mathbf{q}_{[t]}^r+\sum_{i=1}^r(\mathbf{u}_{[t]}^{i}-\mathbf{S}_{[t]}^{0 \top}\mathbf{w}_{[t]}^{i})(\mathbf{k}_{[t]}^{i \top}\mathbf{q}_{[t]}^i)
\end{equation}

letting $\mathbf{S}_{[t]}=\mathbf{S}_{[t]}^0$, the above can be simplified to matrix notations:

\begin{equation}
    \mathbf{S}_{[t+1]} = \mathbf{S}_{[t]} + \mathbf{K}_{[t]}^\top\left( \mathbf{U}_{[t]} - \mathbf{W}_{[t]} \mathbf{S}_{[t]} \right)\end{equation}
\begin{equation}
    \mathbf{O}_{[t]} = \mathbf{Q}_{[t]} \mathbf{S}_{[t]} + \left( \mathbf{Q}_{[t]} \mathbf{K}_{[t]}^\top \odot \mathbf{M} \right) \left( \mathbf{U}_{[t]} - \mathbf{W}_{[t]} \mathbf{S}_{[t]} \right),
\end{equation}
where $\Box_{[t]} = \Box_{[t]}^{1:C} \in \mathbb{R}^{C \times d}$ for 
$\Box \in \{\mathbf{Q}, \mathbf{K}, \mathbf{V}, \mathbf{O}, \mathbf{U}, \mathbf{W}\}$ 
defines the chunkwise matrices that are formed from stacking the 
$\mathbf{q}_t, \mathbf{k}_t, \mathbf{v}_t, \mathbf{o}_t, \mathbf{u}_t, \mathbf{w}_t$ vectors and $\mathbf{M}$ is the lower triangular causal mask.

Finally, we can apply the UT transform~\citep{joffrain2006accumulating} to simplify the recurrence calculations of $\mathbf{u}_{[t]}^r$ and $\mathbf{w}_{[t]}^r$.

\begin{equation}
    \mathbf{T}_{[i]}=
\left(\mathbf{I}+\mathrm{StrictTril}(\mathrm{diag}(\alpha_t)\mathbf{K}_{[i]}\mathbf{K}_{[i]}^\top)\right)^{-1}
\mathrm{diag}(\alpha_t)
\end{equation}

\begin{equation}
\mathbf{W}_{[t]} = \mathbf{T}_{[t]}\mathbf{K}_{[t]}, \qquad
\mathbf{U}_{[t]} = \mathbf{T}_{[t]}\mathbf{V}_{[t]}
\end{equation}

\subsection{Discussion}
\label{sec:memory_dominance}

\textbf{Directional decay of memory.}
The EFLA formulation reveals the role of the key norm in controlling
memory retention. Since
\(\mathbf{A}_t=\mathbf{k}_t\mathbf{k}_t^\top\) is symmetric positive
semi-definite and rank-1, it has one non-zero eigenvalue
\(\lambda_t=\lVert\mathbf{k}_t\rVert^2\) when \(\mathbf{k}_t\neq0\), while all
remaining eigenvalues are zero. Consequently,
\begin{equation}
e^{-\beta_t\mathbf{A}_t}
=
\mathbf{I}
-
\frac{1-e^{-\beta_t\lambda_t}}{\lambda_t}
\mathbf{k}_t\mathbf{k}_t^\top .
\end{equation}
This transition contracts the component of the memory aligned with
\(\mathbf{k}_t\) by the factor \(e^{-\beta_t\lambda_t}\), while leaving the
orthogonal subspace unchanged. For \(\beta_t\ge0\), the eigenvalues of the
homogeneous transition are \(e^{-\beta_t\lambda_t}\) and \(1\), so the
homogeneous flow is non-expansive.

\textbf{Connection to the delta rule.}
The effective update coefficient satisfies:
\begin{equation}
\alpha_t
=
\frac{1-e^{-\beta_t\lambda_t}}{\lambda_t}
=
\beta_t
-
\frac{1}{2}\beta_t^2\lambda_t
+
\mathcal{O}(\beta_t^3\lambda_t^2).
\end{equation}
Therefore, when the effective update scale \(\beta_t\lambda_t\) is small, EFLA
recovers the delta-rule update up to higher-order terms. For larger
\(\beta_t\lambda_t\), EFLA uses the bounded exponential flow, whereas the
Euler-style delta rule keeps only the first-order approximation.

\textbf{Role of key norm and stable dynamics.}
EFLA naturally exposes the role of the key norm through the exact-flow coefficient:
\begin{equation}
\alpha_t=\frac{1-e^{-\beta_t\lambda_t}}{\lambda_t},
\qquad
\lambda_t=\lVert \mathbf{k}_t\rVert^2 .
\end{equation}
The denominator \(\lambda_t\) is not an additional normalization trick, but
arises directly from solving the rank-1 ODE in closed form. This shows that the
strength of the exact-flow update is intrinsically tied to the magnitude of the
current key.

This dependence also affects the write term
\(\alpha_t\mathbf{k}_t\mathbf{v}_t^\top\). When keys are L2-normalized, the
model mainly uses the direction of \(\mathbf{k}_t\). In contrast, EFLA preserves
the magnitude of \(\mathbf{k}_t\) in the key--value injection term, allowing the
current token to modulate the memory write not only through the learned gate
\(\beta_t\), but also through the signal strength encoded in the key norm.

Finally, the continuous-time dynamics provide a stability interpretation. The
homogeneous part of the ODE is governed by:
\begin{equation}
\frac{d\mathbf{S}(\tau)}{d\tau}
=
-\mathbf{k}_t\mathbf{k}_t^\top \mathbf{S}(\tau),
\end{equation}
where \(-\mathbf{k}_t\mathbf{k}_t^\top\) is negative semi-definite. Therefore, the exact transition contracts the memory component aligned with \(\mathbf{k}_t\) by the factor \(e^{-\beta_t\lambda_t}\), while leaving the orthogonal subspace unchanged. This contractive flow suppress perturbations along the current key direction and provides a natural explanation for the improved stability and noise robustness observed in our experiments in Section~\ref{Robustness Study}.

\textbf{Gated Variant of EFLA.}
EFLA is also compatible with gated recurrences by
applying the exact-flow construction to a gated update mechanism. The derivation of this
variant, which we term Gated EFLA, is provided in Appendix~\ref{appendix:efla_gated}.

\section{Experiments}
\label{sec:empirical_study}

\subsection{Experimental Setup}

\textbf{Baselines.}
We compare EFLA and its gated variants with SSM methods including Mamba-2~\citep{dao2024transformers}, and Euler-style methods including
DeltaNet~\citep{yang2024parallelizing}, and Gated DeltaNet~\citep{yang2024gated}.
EFLA is implemented as a drop-in replacement for the delta-rule
state update in Euler-style methods, using the exact-flow coefficient derived in
Section~\ref{sec:efla}. 

\begin{table}[htbp]
\centering
\setlength{\tabcolsep}{4pt}
\caption{Main language modeling results.
The 340M models are trained for 8B tokens, whereas the 1.3B models are trained for 50B tokens on the same subset of the SlimPajama dataset~\citep{cerebras2023slimpajama} with the Mistral~\citep{jiang2023mistral7b} tokenizer. \textbf{Perplexity:} Lower ($\downarrow$) is better. \textbf{Accuracy:} Higher ($\uparrow$) is better. Best results are bolded.}
\resizebox{0.95\linewidth}{!}{%
\begin{tabular}{l cc ccccccccc c}
\toprule
& \multicolumn{2}{c}{\textbf{Perplexity} ($\downarrow$)} & \multicolumn{9}{c}{\textbf{Accuracy} ($\uparrow$)} & \\
\cmidrule(lr){2-3} \cmidrule(lr){4-12}
\textbf{Model} & 
\textbf{Wiki.} & \textbf{LMB.} & 
\textbf{LMB.} & \textbf{PIQA} & \textbf{Hella.} & \textbf{Wino.} & \textbf{ARC-e} & \textbf{ARC-c} & \textbf{BoolQ} & \textbf{OBQA} & \textbf{SciQ} & 
\textbf{Avg.} \\
& \scriptsize{ppl} & \scriptsize{ppl} & \scriptsize{acc} & \scriptsize{acc} & \scriptsize{acc\_n} & \scriptsize{acc} & \scriptsize{acc} & \scriptsize{acc\_n} & \scriptsize{acc}  & \scriptsize{acc\_n} & \scriptsize{acc} & \\
\midrule

\multicolumn{13}{l}{\textit{\textbf{340M Parameters}}} \\
\midrule
\rowcolor{gray!15}
\multicolumn{13}{l}{\textbf{Vanilla}} \\
Mamba-2 & 36.90 & 105.69 & 20.7 & \textbf{61.9} & 30.7 & 50.2 & 40.4 & 22.1 & 53.0 & 27.0 & 71.9 & 42.0 \\
DeltaNet & 38.09 & 96.26 & 22.5 & 60.7 & 30.1 & \textbf{51.9} & 41.7 & 21.6 & 60.1 & 29.0 & 70.4 & 42.2 \\
EFLA & 35.26 & \textbf{79.97} & \textbf{23.9} & 61.0 & 30.9 & 51.1 & \textbf{42.5} & \textbf{23.8} & 59.7 & \textbf{30.8} & 72.9 & \textbf{44.1} \\

\rowcolor{gray!15}
\multicolumn{13}{l}{\textbf{Gated Variants}} \\

Gated DeltaNet & 34.93 & 72.46 & 25.5 & \textbf{62.7} & 30.8 & 49.3 & 41.8 & 21.7 & 61.7 & 27.4 & 74.2 & 43.9 \\
Gated EFLA & \textbf{34.28} & \textbf{69.37} & \textbf{26.4} & 61.8 & \textbf{31.1} & \textbf{52.1} & \textbf{42.0} & \textbf{22.8} & \textbf{62.1} & \textbf{29.4} & \textbf{75.3} & \textbf{44.8} \\

\midrule

\multicolumn{13}{l}{\textit{\textbf{1.3B Parameters}}} \\
\midrule
DeltaNet & 18.38 & 17.29 & 41.8 & \textbf{69.2} & \textbf{44.5} & 49.3 & 52.5 & \textbf{26.4} & 58.1  & 29.8 & 82.6 & 50.5 \\
EFLA     & \textbf{18.30} & \textbf{16.54} & \textbf{43.2} & 68.9 & \textbf{44.5} & \textbf{52.1} & \textbf{54.4} & \textbf{26.4} & \textbf{60.4}  & \textbf{31.6} & \textbf{84.2} & \textbf{51.8} \\

\bottomrule
\end{tabular}
}
\label{tab:full_comparison}
\end{table}

\textbf{Training setup.}
The 340M models are trained for 8B tokens, and the 1.3B models are trained for
50B tokens on the same subset of the SlimPajama dataset~\citep{cerebras2023slimpajama}
with the Mistral tokenizer~\citep{jiang2023mistral7b}. Detailed hyperparameter
settings are provided in Appendix~\ref{appendix:experimental_setting}.

\textbf{Evaluation.}
We evaluate language modeling perplexity on Wikitext~\citep{merity2016pointer}
and LAMBADA~\citep{paperno2016lambada}, together with a suite of zero-shot
commonsense reasoning tasks, including PiQA~\citep{bisk2020piqa},
HellaSwag~\citep{zellers2019hellaswag}, WinoGrande~\citep{sakaguchi2021winogrande},
ARC-easy (ARC-e), ARC-challenge (ARC-c)~\citep{clark2018think},
BoolQ~\citep{clark2019boolq}, OpenBookQA (OBQA)~\citep{OpenBookQA2018},
and SciQ~\citep{SciQ}. We report perplexity for Wikitext and LAMBADA, and
accuracy for the downstream reasoning tasks.

\subsection{Main Language Modeling Results}

The main language modeling results are shown in Table~\ref{tab:full_comparison}.
For 340M-parameter models trained with the same 8B-token budget, EFLA consistently improves over the Euler-style baseline on most metrics. On the perplexity metrics, EFLA reduces Wikitext perplexity from 38.09 to 35.26 and LAMBADA perplexity from 96.26 to 79.97, indicating better language modeling quality under the same training budget. On downstream accuracy, EFLA improves the average score from 42.2\% to 44.1\%, suggesting that the exact-flow update improves the general effectiveness of the Euler-style recurrence. The comparison with Mamba-2 further shows that EFLA is competitive among efficient recurrent architectures at this scale.

The same trend holds for the gated variants. Replacing the Gated DeltaNet update with the Gated EFLA update reduces Wikitext perplexity from 34.93 to 34.28 and LAMBADA perplexity from 72.46 to 69.37. It also improves the average downstream accuracy from 43.9\% to 44.8\%. This result is important because it shows that the exact-flow update is not limited to the vanilla formulation, but can also be incorporated into the gated variant and still produce consistent improvements.

At the larger 1.3B scale, the improvements persist under a longer 50B-token training budget. Compared with DeltaNet, EFLA reduces Wikitext perplexity from 18.38 to 18.30 and LAMBADA perplexity from 17.29 to 16.54. It also improves the average downstream accuracy from 50.5\% to 51.8\%. Therefore, the improvement is not a small-scale artifact, the exact-flow update continues to provide benefits when both model size and training tokens are increased.

\subsection{Robustness under Corrupted and High-energy Inputs}
\label{sec:robustness_study}

\textbf{Robustness evaluation.}
\label{Robustness Study}
As discussed in Section~\ref{sec:efla}, Euler discretization delta-rule updates can be interpreted as explicit Euler discretizations of the corresponding ZOH
continuous-time dynamics. While computationally efficient, this first-order discretization may become inaccurate under large effective update scales or corrupted inputs. To evaluate this behavior, we conduct three robustness tests on pixel-level Sequential MNIST~\citep{lecun2010mnist}.

\begin{figure}[htbp]
    \centering
    \includegraphics[width=\linewidth]{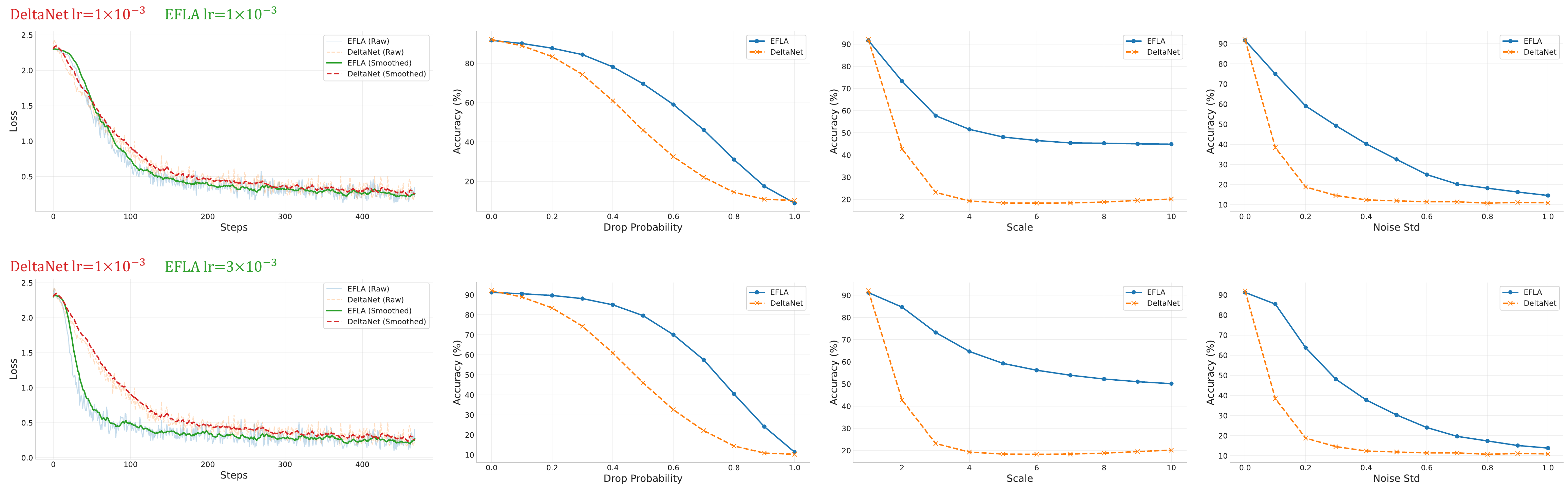}
    \caption{EFLA outperforms DeltaNet in both convergence speed and robustness on sMNIST. The plots illustrate training dynamics and robustness against dropout, intensity scaling, and additive Gaussian noise. EFLA maintains higher accuracy as perturbation intensity increases, especially under the larger learning-rate setting \(3\times10^{-3}\).}
    \label{fig:smnist}
\end{figure}

\textbf{Perturbation settings.}
We consider three types of input perturbations. First, \textbf{image pixel
dropout} applies Bernoulli dropout to input tokens with probability \(p\),
simulating information loss in corrupted inputs. Second, \textbf{OOD intensity
scaling} amplifies input signals by a fixed factor, testing stability under
high-energy inputs. Third, \textbf{additive Gaussian noise} injects random noise
with varying standard deviations, evaluating robustness to signal corruption.

\textbf{Experimental setup.}
We flatten each \(28\times28\) image into a sequence of length \(L=784\) and use
a linear attention classifier with hidden dimension \(d=64\). We compare EFLA
with the DeltaNet baseline. Both models are trained with AdamW using a batch
size of 128.

\textbf{Results.}
Figure~\ref{fig:smnist} shows that DeltaNet is more sensitive to input scaling, with accuracy dropping rapidly as the scale factor increases. In contrast, EFLA maintains higher accuracy under large input scales, consistent with its exact-flow transition, whose homogeneous component remains non-expansive for non-negative decay. EFLA also outperforms DeltaNet under additive noise and pixel dropout, with a slower degradation rate under stronger perturbations. These results support the view that removing the Euler discretization error of the ZOH delta-rule state transition improves robustness under corrupted or high-energy inputs.

\subsection{Robustness to Learning Rate Selection}
\label{sec:lr-robustness}
To further examine the optimization behavior of EFLA, we evaluate its sensitivity to the learning rate on the Sequential MNIST. We compare EFLA with DeltaNet under the same training setup while varying the learning rate over $\{1\times10^{-4}, 3\times10^{-4}, 10^{-3}, 3\times10^{-3}, 1\times10^{-2}\}$.

\begin{figure}[t]
    \centering
    \begin{subfigure}[b]{0.45\linewidth}
        \centering
        \includegraphics[width=\linewidth]{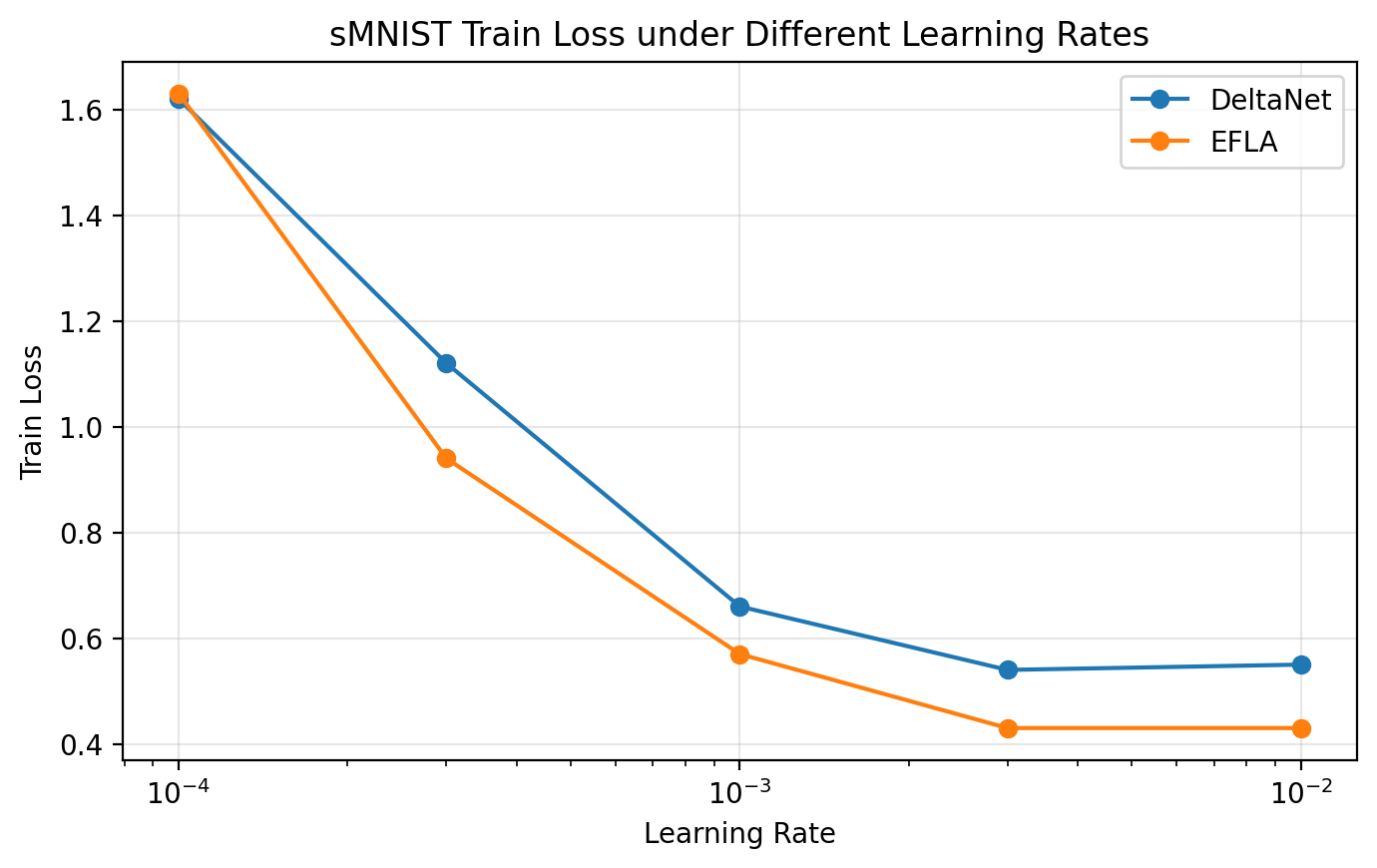}
        \caption{Training loss.}
        \label{fig:smnist_lr_loss}
    \end{subfigure}
    \hfill
    \begin{subfigure}[b]{0.45\linewidth}
        \centering
        \includegraphics[width=\linewidth]{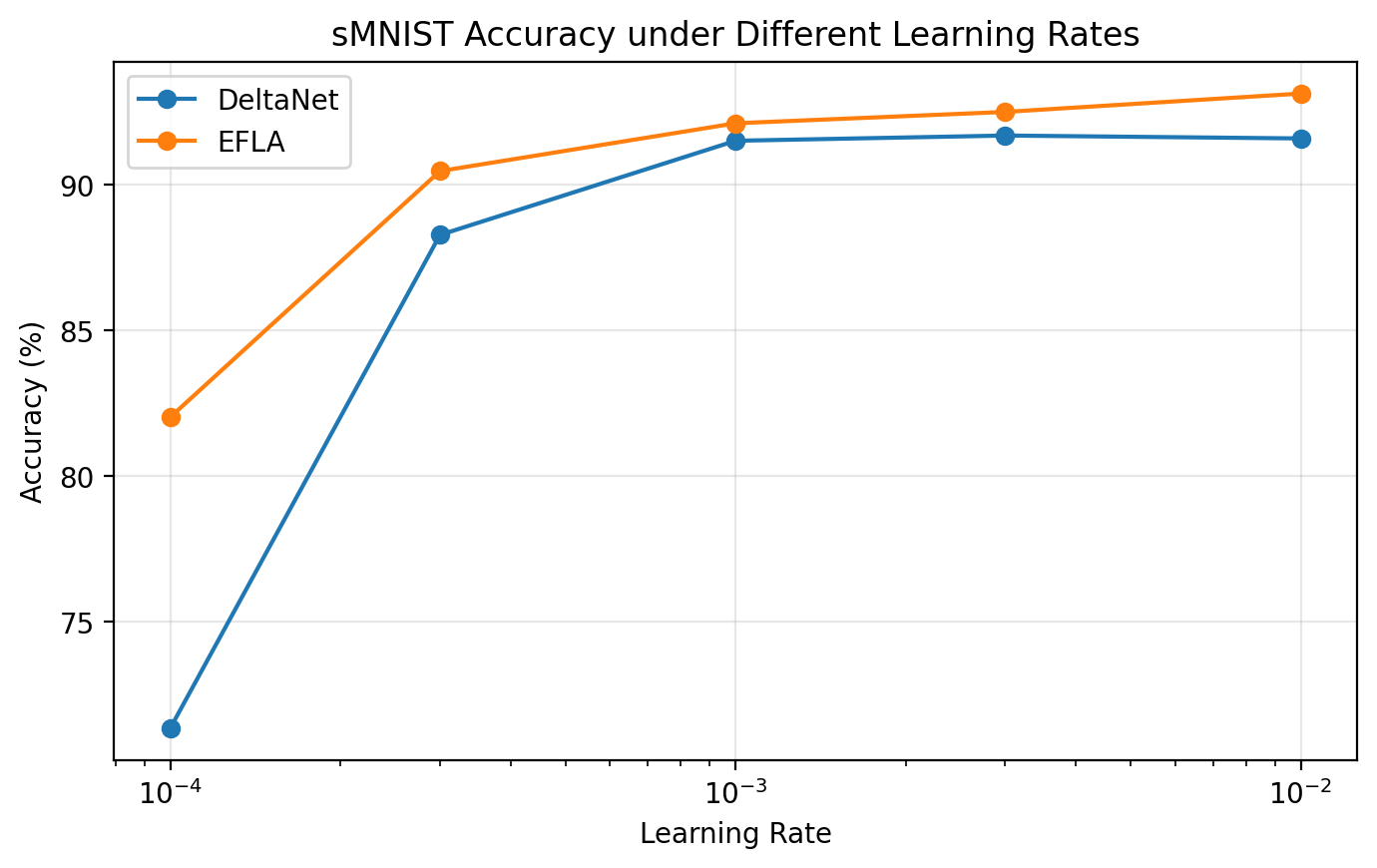}
        \caption{Accuracy.}
        \label{fig:smnist_lr_acc}
    \end{subfigure}
    \caption{sMNIST results under different learning rates. EFLA maintains lower training loss and higher accuracy across the tested learning-rate range, indicating stronger robustness to learning-rate selection.}
    \label{fig:smnist_lr_robustness}
\end{figure}

As the results shown in Figure~\ref{fig:smnist_lr_robustness}, EFLA remains competitive across the entire learning-rate range. At very small learning rates, such as $1\times10^{-4}$, both methods have relatively high training loss, but EFLA already achieves substantially higher accuracy than DeltaNet. As the learning rate increases, EFLA consistently obtains lower training loss and higher accuracy. Notably, when the learning rate is increased to $1\times10^{-2}$, DeltaNet shows no further accuracy improvement, whereas EFLA continues to achieve the best accuracy among all tested settings. These results suggest that EFLA is less sensitive to learning-rate selection than the Euler-style DeltaNet update. This behavior is consistent with our theoretical interpretation: by reducing the discretization error introduced at each recurrent state transition, EFLA provides a more faithful state update, which can improve optimization behavior across different learning-rate regimes.

\subsection{Results on the MAD Synthetic Benchmark}


\begin{figure}[htbp]
    \centering
    \begin{minipage}[t]{0.48\textwidth}
        \centering
        \includegraphics[width=\linewidth]{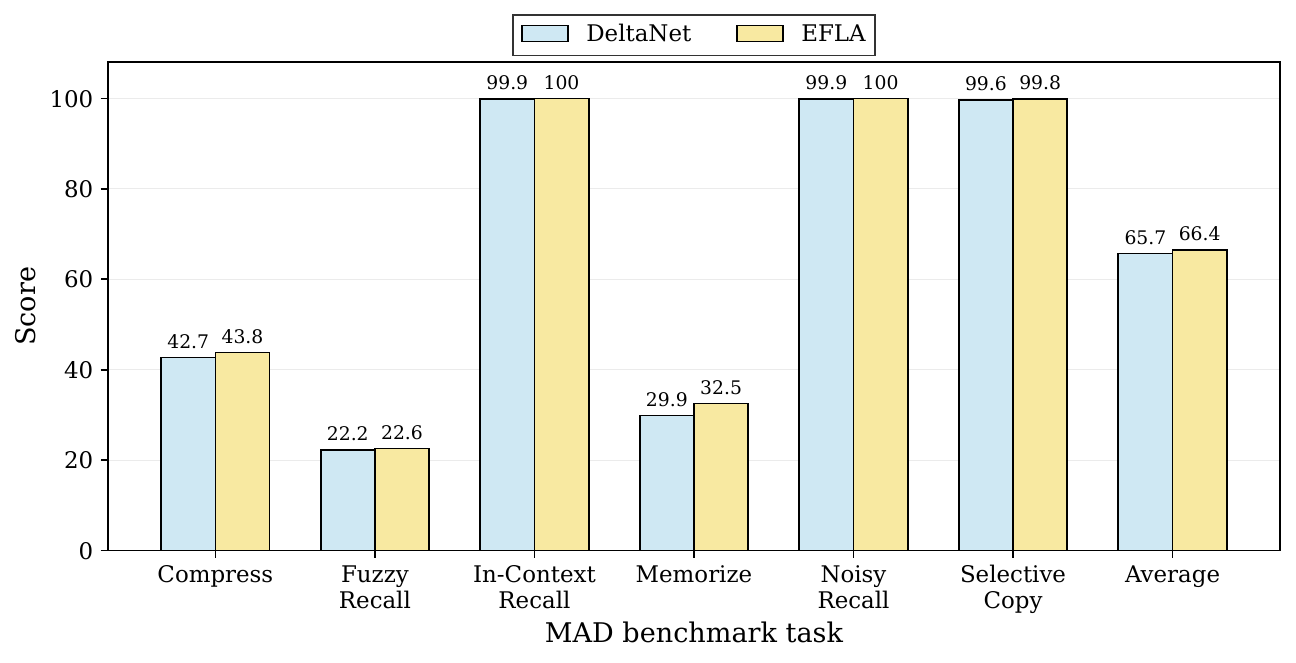}
        \caption{Results on the synthetic MAD benchmark.}
        \label{fig:mad_benchmark}
    \end{minipage}
    \hfill
    \begin{minipage}[t]{0.44\textwidth}
        \centering
        \includegraphics[width=\linewidth]{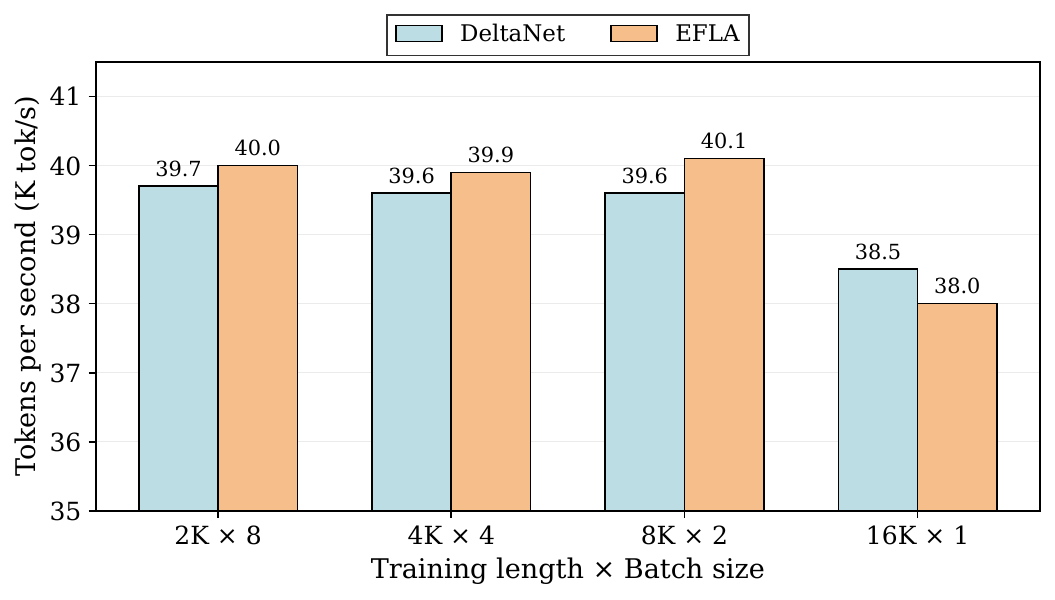}
        \caption{Training throughput of DeltaNet and EFLA.}
        \label{fig:throughput_comparison}
    \end{minipage}
\end{figure}

We further evaluate EFLA on the Mechanistic Architecture Design (MAD)
benchmark~\citep{poli2024mechanisticdesignscalinghybrid}, a suite of synthetic token-manipulation tasks designed to isolate architecture-level capabilities. As shown in Figure~\ref{fig:mad_benchmark}, EFLA improves upon DeltaNet across all six tasks, increasing the average score from 65.7\% to 66.4\%. The gains are most visible on Memorize and Compress, suggesting that the exact-flow state transition improves the model's ability to maintain and update token-level memory.

\subsection{Training Efficiency}

We compare the training efficiency of DeltaNet and EFLA under matched
sequence-length and batch-size configurations. As shown in
Figure~\ref{fig:throughput_comparison}, EFLA maintains nearly identical
throughput to DeltaNet across all settings. These results confirm that the
exact-flow update preserves the hardware-efficient structure of the delta-rule
recurrence and does not introduce meaningful training overhead.

\section{Related Work}
\label{sec:related_work}

\textbf{Kernel Linear Attention and Delta-Rule Updates.}
Linear-time attention methods reduce the quadratic cost of softmax attention by
replacing or reformulating the softmax operation. Kernel-based methods such as
Linear Transformers~\citep{katharopoulos2020transformers} and
Performer~\citep{choromanski2020rethinking} express causal attention through
recurrently accumulated key--value statistics. From the fast-weight perspective,
\citet{schlag2021linear} interpret this matrix-valued state as an associative
memory.

Delta-rule linear attention~\citep{schlag2021linear,yang2024parallelizing}
improves this memory update by solving an online reconstruction problem. A
gradient step with step size \(\beta_t\) gives
\begin{equation}
\mathbf{S}_t
=
(\mathbf{I}-\beta_t\mathbf{k}_t\mathbf{k}_t^\top)\mathbf{S}_{t-1}
+
\beta_t\mathbf{k}_t\mathbf{v}_t^\top .
\end{equation}
This update introduces input-dependent correction and forgetting. Later variants
such as Gated DeltaNet~\citep{yang2024gated} and Kimi Delta Attention
(KDA)~\citep{team2025kimi} further strengthen Euler discretization recurrences through
gating, adaptive update coefficients, and richer value mixing. In contrast, our
work revisits the same delta-rule dynamics from a numerical-integration
perspective: under a zero-order-hold (ZOH) formulation, the delta-rule update is
an explicit Euler discretization. EFLA replaces this Euler transition with the
exact ZOH flow induced by the rank-1 matrix
\(\mathbf{A}_t=\mathbf{k}_t\mathbf{k}_t^\top\), while preserving the
computational structure.

\textbf{State Space Models and Continuous-Time Discretization.}
State Space Models (SSMs) provide another route to efficient long-sequence
modeling by deriving discrete-time sequence operators from continuous-time
dynamical systems. The S4 family~\citep{gu2022s4,gu2022s4d,gu2020hipporecurrentmemoryoptimal}
uses structured state matrices and discretization schemes such as the bilinear
transform and ZOH. Mamba~\citep{gu2024mamba} introduces input-dependent
selectivity with a hardware-efficient scan, and Mamba-2~\citep{dao2024transformers}
connects selective SSMs with gated linear attention. EMA-based models~\citep{fu2022hungry,ma2022mega,sun2023retentive}
can also be viewed as scalar or diagonal state-space recurrences.

EFLA is related to SSMs in that it also starts from a continuous-time ODE and
derives a discrete-time update. The key difference is the transition structure:
SSMs typically rely on scalar, diagonal, or specially structured matrices,
whereas delta-rule linear attention induces a token-dependent rank-1 transition
matrix. This rank-1 structure makes both the matrix exponential and input
integral analytically tractable, yielding an exact token-wise ZOH flow that
retains the algebraic form needed for efficient WY/UT-style chunkwise
parallelization.

\vspace{-0.8em}
\section{Conclusion}
\label{sec:conclusion}

We presented \textbf{Exact Flow Linear Attention (EFLA)}, an exact-flow
formulation of delta-rule linear attention. By interpreting the Euler-style Delta-rule
state update as an explicit Euler discretization of a zero-order-hold (ZOH)
continuous-time system, we identified a principled source of approximation error
in the standard delta-rule recurrence. Instead of heuristically modifying this
Euler-style update, EFLA derives the closed-form flow of the underlying ZOH
dynamics. By exploiting the token-wise rank-1 transition matrix
\(\mathbf{A}_t=\mathbf{k}_t\mathbf{k}_t^\top\), the resulting update removes the
Euler discretization error while preserving the algebraic structure, parameter
count, and linear-time complexity of DeltaNet.
Empirically, we showed that this exact-flow replacement leads to consistent
gains across robustness tests, language modeling benchmarks, and the MAD
synthetic benchmark. EFLA improves robustness under corrupted and high-energy
inputs, achieves lower perplexity, and obtains stronger downstream performance
than DeltaNet without introducing additional parameters. Moreover, its
compatibility with WY/UT-style chunkwise parallelization preserves practical
training efficiency. These results suggest that exact-flow integration provides
a principled and scalable way to improve delta-rule linear attention, and may
inspire future work on exact solvers for more general continuous-time
attention-like architectures.

\section*{Acknowledgment}
This research/project is supported by the National Research Foundation, Singapore under its AI
Singapore Programme (AISG Award No: AISG3-GV2023-010). This work is also supported by
the National Research Foundation, Singapore, under its National Large Language Models Funding
Initiative (AISG Award No: AISG-NMLP-2024-005), and NTU SUG project \#025628-00001: Posttraining to Improve Embodied AI Agents. This research
is supported by the National Research Foundation, Prime
Minister’s Office, Singapore under its Campus for Research
Excellence and Technological Enterprise (CREATE) programme.

\bibliographystyle{plainnat}
\bibliography{example_paper}

\newpage
\appendix

\section{Limitations}
\label{appendix:limitations}

EFLA is derived under a zero-order-hold (ZOH) continuous-time formulation of Euler-style delta-rule dynamics. This formulation enables an exact closed-form flow by exploiting the token-wise rank-1 transition matrix \(\mathbf{A}_t=\mathbf{k}_t\mathbf{k}_t^\top\), and directly removes the Euler discretization error of the corresponding ZOH state transition. However, this does not eliminate all sources of error in practical sequence modeling. Models may still be affected by finite-precision computation, optimization dynamics, data limitations, architectural choices, and the approximation gap between linear attention and softmax attention. A more complete analysis of how exact-flow updates interact with these factors is left for future work.

Our experiments validate EFLA through controlled robustness tests, language modeling at 340M and 1.3B scales, downstream benchmarks, the MAD synthetic benchmark, and training-throughput comparisons. Nevertheless, due to the substantial computational cost of training larger long-context models from scratch, we do not yet provide exhaustive scaling experiments beyond the 1.3B regime or extensive evaluations on deployment-oriented long-context tasks. Although EFLA preserves the linear-time recurrence and WY/UT-compatible chunkwise-parallel structure of DeltaNet, further empirical characterization is needed to understand its scaling behavior under larger pretraining budgets, longer context windows, and broader downstream settings.

\section{Broader Impacts}
\label{appendix:broad_impact}
This work proposes a methodological improvement to linear-time attention by deriving an exact-flow update for delta-rule linear attention. Its potential positive impact lies in improving the efficiency and robustness of long-sequence modeling, which may reduce the computational cost and energy consumption of training and deploying sequence models. More efficient recurrent attention mechanisms may also make long-context processing more accessible under limited hardware budgets.

At the same time, efficiency improvements can lower the cost of deploying large language models more broadly. As with other advances in efficient sequence modeling, this may indirectly amplify both beneficial applications and potential misuse, including unsafe generation, automated misinformation, or other harmful uses of language models. This paper does not introduce a new dataset, release a deployed model, or target a specific high-risk application domain. Therefore, we do not study application-specific societal effects, and any deployment of models using EFLA should follow standard safety evaluation and responsible-release practices.

\section{Experimental Setting}
\label{appendix:experimental_setting}
We used 8 A100$\times$80G GPUs for 340M and 1.3B language modeling experiments. The random seed it set to 42. Each model uses AdamW for optimization, with a peak learning rate of $3 \times 10^{-4}$. The 340M models are trained for 8 billion tokens with a global batch size of 1M tokens, while the 1.3B models are trained for 50 billion tokens with a global batch size of 2M tokens. We use a cosine learning rate schedule, starting with a warm-up phase of 1 billion tokens for the 340M models and 2 billion tokens for the 1.3B models (corresponding to 1024 warm-up steps). Both have configurations that initial and final learning rates set at $3 \times 10^{-5}$. We apply a weight decay of 0.1 and use gradient clipping at a maximum of 1.0. The head dimension is set to 128, and the kernel size for convolution layers is set at 4. To ensure numerical stability, specifically to prevent division by zero when the key norm $\|\mathbf{k}_t\|^2$ vanishes, we clip it with a lower bound of $\epsilon = 1\times10^{-12}$. Additionally, we employ the \textit{expm1} function to compute the numerator $1-e^{-\beta_t\lambda_t}$, preserving precision for small exponents. We empirically find that directly using the projected keys, or applying \(L_2\) normalization to the keys, leads to performance similar to DeltaNet. In contrast, scaling the keys by \(\sqrt{d}\), where \(d\) is the head dimension, consistently gives better results. A possible explanation is that the difference between the exact-flow update becomes more pronounced when the effective key magnitude is larger, whereas small or normalized keys tend to place the model in a regime where the two updates behave similarly. This observation is also consistent with the role of the key norm in the exact-flow formulation, where the key magnitude directly affects the state transition and memory update. Importantly, such scaling is naturally supported by EFLA whereas applying the same scaling to the Euler-style DeltaNet update can lead to overly aggressive or unstable state transitions.

\section{Construction and Properties of \textit{rank-1} Matrices}
\label{appendix:property-rank1-matrices}

$\mathbf{A}_t$ is a \textit{rank-1} matrix, and it satisfies:
\begin{equation}
\mathbf{A}_t^2 = \mathbf{k}_t \mathbf{k}_t^\top \mathbf{k}_t \mathbf{k}_t^\top = \mathbf{k}_t (\mathbf{k}_t^\top \mathbf{k}_t) \mathbf{k}_t^\top = \lambda_t \mathbf{A}_t,
\end{equation}

Where $\lambda_t = \mathbf{k}_t^\top \mathbf{k}_t$ is scalar value.

Then it gives us a key property: \( \mathbf{A}_t \) is a scaled projection matrix (i.e., \( \mathbf{A}_t^2 = \lambda_t \mathbf{A}_t \)).

\section{General Solutions of Ordinary Differential Equations}
\label{appendix:general-solution-ode}

We start with a first-order linear matrix ODE:
\begin{equation}
\frac{d\mathbf{S}}{dt} = -A\mathbf{S}+\mathbf{b},
\end{equation}
Which can be rewrite as:
\begin{equation}
\frac{d\mathbf{S}}{dt} + A\mathbf{S} = \mathbf{b},
\end{equation}

For this type of differential equation, the integrating factor is:
\begin{equation}
e^{\int\mathbf{A} \, dt},
\end{equation}
Since \( \mathbf{A} \) is constant, the integrating factor is simply:
\begin{equation}
e^{\mathbf{A}t}
\end{equation}

Multiply the entire equation by \( e^{\mathbf{A}t} \):
\begin{equation}
e^{\mathbf{A}t} \left( \frac{d\mathbf{S}}{dt} + A\mathbf{S} \right) = e^{\mathbf{A}t} \mathbf{b},
\end{equation}

Expanding the left-hand side:
\begin{equation}
\label{eq:ode_3}
e^{\mathbf{A}t} \frac{d\mathbf{S}}{dt} + e^{\mathbf{A}t} \mathbf{A} \mathbf{S} = e^{\mathbf{A}t} \mathbf{b},
\end{equation}

By the product rule for matrix-vector multiplication:
\begin{equation}
\frac{d}{dt}(e^{\mathbf{A}t} \mathbf{S}) = \left( \frac{d}{dt} e^{\mathbf{A}t} \right) \mathbf{S} + e^{\mathbf{\mathbf{A}}t} \frac{d\mathbf{S}}{dt},
\end{equation}
and since \( \frac{d}{dt} e^{\mathbf{A}t} = \mathbf{A} e^{\mathbf{A}t} \), we have:
\begin{equation}
\frac{d}{dt}(e^{\mathbf{A}t} \mathbf{S}) = (\mathbf{A} e^{\mathbf{A}t})\mathbf{S} + e^{\mathbf{A}t}\frac{d\mathbf{S}}{dt},
\end{equation}
Notice this matches exactly the left-hand side of Eq.~\ref{eq:ode_3} (since \(\mathbf{A}\) and \(e^{\mathbf{A}t}\) commute).

The equation becomes:
\begin{equation}
\frac{d}{dt}(e^{\mathbf{A}t} \mathbf{S}) = e^{\mathbf{A}t} \mathbf{b},
\end{equation}
Integrate both sides from \( t \) (initial time) to \( t+\beta_t \) (final time).  
To avoid confusion, we use \( \tau \) as the integration variable:
\begin{equation}
\int_{t}^{t+\beta_t} \frac{d}{d\tau}(e^{A\tau} \mathbf{S}(\tau)) \, d\tau = \int_{t}^{t+\beta_t} e^{A\tau} \mathbf{b} \, d\tau,
\end{equation}

\begin{equation}
\left[ e^{\mathbf{A}\tau} \mathbf{S}(\tau) \right]_{t}^{t+\beta_t} = \int_{t}^{t+\beta_t} e^{\mathbf{A}\tau} \mathbf{b} \, d\tau,
\end{equation}
Thus:
\begin{equation}
e^{\mathbf{A}(t+\beta_t)} \mathbf{S}(t+\beta_t) - e^{\mathbf{A}t} \mathbf{S}(t) = \int_{t}^{t+\beta_t} e^{\mathbf{A}\tau} \mathbf{b} \, d\tau,
\end{equation}

Multiply both sides by \( e^{-\mathbf{A}(t+\beta_t)} \):
\begin{equation}
\mathbf{S}(t+\beta_t) - e^{-\mathbf{A}(t+\beta_t)} e^{\mathbf{A}t} \mathbf{S}(t) = e^{-\mathbf{A}(t+\beta_t)} \int_{t}^{t+\beta_t} e^{\mathbf{A}\tau} \mathbf{b} \, d\tau,
\end{equation}

Simplify using exponential properties:
\begin{equation}
\mathbf{S}(t+\beta_t) - e^{-\mathbf{A}\beta_t} \mathbf{S}(t) = \int_{t}^{t+\beta_t} e^{-\mathbf{A}(t+\beta_t-\tau)} \mathbf{b} \, d\tau,
\end{equation}

Since \( e^{-\mathbf{A}(t+\beta_t)} \) is constant, it can be moved inside the integral.

Let \( s = \tau - t \).  
Then:
\begin{equation}
\begin{cases}
s = 0, & \tau = t \\
s = \beta_t, & \tau = t+\beta_t
\end{cases}
\quad \text{and} \quad d\tau = ds,
\end{equation}

Substitute:
\begin{equation}
\int_{0}^{\beta_t} e^{-\mathbf{A}[t+\beta_t-(s+t)]} \mathbf{b} \, ds = \int_{0}^{\beta_t} e^{-\mathbf{A}(\beta_t-s)} \mathbf{b} \, ds,
\end{equation}

Rename \( s \) back to \( \tau \) (dummy variable) and replace constants with \( \mathbf{A}_t, \mathbf{b}_t \):
\begin{equation}
\int_0^{\beta_t} e^{-(\beta_t - \tau)\mathbf{A}_t} \mathbf{b}_t \, d\tau,
\end{equation}

Combining everything, the full solution is:
\begin{equation}
\mathbf{S}(t+\beta_t) = e^{-\beta_t \mathbf{A}_t} \mathbf{S}(t)
+ \int_0^{\beta_t} e^{-(\beta_t - \tau)\mathbf{A}_t} \mathbf{b}_t \, d\tau,
\end{equation}

\section{Exact-flow formulation of Gated DeltaNet.}
\label{appendix:efla_gated}
Gated DeltaNet updates the fast-weight state by
\begin{equation}
\mathbf{S}_t
=
\alpha_t
\left(
\mathbf{I}
-
\beta_t\mathbf{k}_t\mathbf{k}_t^\top
\right)
\mathbf{S}_{t-1}
+
\beta_t\mathbf{k}_t\mathbf{v}_t^\top .
\end{equation}

Expanding the transition term gives
\begin{equation}
\alpha_t
\left(
\mathbf{I}
-
\beta_t\mathbf{k}_t\mathbf{k}_t^\top
\right)
=
\mathbf{I}
-
\left[
(1-\alpha_t)\mathbf{I}
+
\alpha_t\beta_t\mathbf{k}_t\mathbf{k}_t^\top
\right].
\end{equation}
Therefore, the update can be written in Euler form as
\begin{equation}
\mathbf{S}_t
=
\mathbf{S}_{t-1}
-
\left[
(1-\alpha_t)\mathbf{I}
+
\alpha_t\beta_t\mathbf{k}_t\mathbf{k}_t^\top
\right]
\mathbf{S}_{t-1}
+
\beta_t\mathbf{k}_t\mathbf{v}_t^\top .
\end{equation}
Thus, it can be interpreted as an explicit Euler discretization of
\begin{equation}
\frac{d\mathbf{S}(\tau)}{d\tau}
=
-
\left[
(1-\alpha_t)\mathbf{I}
+
\alpha_t\beta_t\mathbf{k}_t\mathbf{k}_t^\top
\right]
\mathbf{S}(\tau)
+
\beta_t\mathbf{k}_t\mathbf{v}_t^\top .
\end{equation}

Let
\begin{equation}
\delta_t=1-\alpha_t,\qquad
\rho_t=\alpha_t\beta_t,\qquad
\lambda_t=\mathbf{k}_t^\top\mathbf{k}_t,
\end{equation}
and define
\begin{equation}
\mathbf{A}_t
=
\delta_t\mathbf{I}
+
\rho_t\mathbf{k}_t\mathbf{k}_t^\top .
\end{equation}
The ODE becomes
\begin{equation}
\frac{d\mathbf{S}(\tau)}{d\tau}
=
-\mathbf{A}_t\mathbf{S}(\tau)
+
\beta_t\mathbf{k}_t\mathbf{v}_t^\top .
\end{equation}
Under the zero-order hold assumption, the exact solution over one update
interval is
\begin{equation}
\mathbf{S}_t
=
e^{-\mathbf{A}_t}\mathbf{S}_{t-1}
+
\int_0^1
e^{-(1-\tau)\mathbf{A}_t}
\beta_t\mathbf{k}_t\mathbf{v}_t^\top
\,d\tau .
\end{equation}

For the transition term, since:
\begin{equation}
\mathbf{A}_t
=
\delta_t\mathbf{I}
+
\rho_t\mathbf{k}_t\mathbf{k}_t^\top ,
\end{equation}
we have:
\begin{equation}
e^{-\mathbf{A}_t}
=
e^{-\delta_t}
e^{-\rho_t\mathbf{k}_t\mathbf{k}_t^\top}.
\end{equation}
Using the rank-one matrix exponential identity:
\begin{equation}
e^{-\rho_t\mathbf{k}_t\mathbf{k}_t^\top}
=
\mathbf{I}
-
\frac{1-e^{-\rho_t\lambda_t}}{\lambda_t}
\mathbf{k}_t\mathbf{k}_t^\top ,
\end{equation}
we obtain:
\begin{equation}
e^{-\mathbf{A}_t}\mathbf{S}_{t-1}
=
e^{-\delta_t}
\left[
\mathbf{I}
-
\frac{1-e^{-\rho_t\lambda_t}}{\lambda_t}
\mathbf{k}_t\mathbf{k}_t^\top
\right]
\mathbf{S}_{t-1}.
\end{equation}

For the input term, since:
\begin{equation}
\mathbf{A}_t\mathbf{k}_t
=
(\delta_t+\rho_t\lambda_t)\mathbf{k}_t,
\end{equation}
we have:
\begin{equation}
e^{-s\mathbf{A}_t}\mathbf{k}_t\mathbf{v}_t^\top
=
e^{-s(\delta_t+\rho_t\lambda_t)}
\mathbf{k}_t\mathbf{v}_t^\top .
\end{equation}
Let
\begin{equation}
\eta_t
=
\delta_t+\rho_t\lambda_t
=
(1-\alpha_t)+\alpha_t\beta_t\lambda_t .
\end{equation}
Then:
\begin{align}
\int_0^1
e^{-(1-\tau)\mathbf{A}_t}
\beta_t\mathbf{k}_t\mathbf{v}_t^\top
\,d\tau
&=
\beta_t
\int_0^1
e^{-(1-\tau)\eta_t}
\,d\tau\,
\mathbf{k}_t\mathbf{v}_t^\top
\\
&=
\beta_t
\frac{1-e^{-\eta_t}}{\eta_t}
\mathbf{k}_t\mathbf{v}_t^\top .
\end{align}

Combining the transition and input terms gives the exact-flow counterpart:
\begin{equation}
\mathbf{S}_t
=
e^{-(1-\alpha_t)}
\left[
\mathbf{I}
-
\frac{1-e^{-\alpha_t\beta_t\lambda_t}}{\lambda_t}
\mathbf{k}_t\mathbf{k}_t^\top
\right]
\mathbf{S}_{t-1}
+
\beta_t
\frac{
1-e^{-[(1-\alpha_t)+\alpha_t\beta_t\lambda_t]}
}{
(1-\alpha_t)+\alpha_t\beta_t\lambda_t
}
\mathbf{k}_t\mathbf{v}_t^\top,
\end{equation}
where $\lambda_t=\mathbf{k}_t^\top\mathbf{k}_t$.

\end{document}